\documentclass[10pt,twocolumn,letterpaper]{article}

\usepackage{cvpr}

\usepackage{graphicx}
\usepackage{amsmath}
\usepackage{amssymb}
\usepackage{booktabs}
\usepackage{multirow} 
\usepackage[pagebackref,breaklinks,colorlinks]{hyperref}
\usepackage{wasysym}
\usepackage{pifont}
\usepackage[accsupp]{axessibility}  % Improves PDF readability for those with disabilities.

\usepackage[capitalize]{cleveref}
\crefname{section}{Sec.}{Secs.}
\Crefname{section}{Section}{Sections}
\Crefname{table}{Table}{Tables}
\crefname{table}{Tab.}{Tabs.}

\def\cvprPaperID{108} % *** Enter the CVPR Paper ID here
\def\confName{CVPR}
\def\confYear{2023}

\begin{document}
	
	%%%%%%%%% TITLE - PLEASE UPDATE
	\title{SCANet: Self-Paced Semi-Curricular Attention Network for \\ Non-Homogeneous Image Dehazing}
	
	\author{
        Yu Guo\textsuperscript{1} ~~ 
        Yuan Gao\textsuperscript{1} ~~ 
        Ryan Wen Liu\textsuperscript{1}\thanks{Corresponding author (wenliu@whut.edu.cn).} ~~ 
        Yuxu Lu\textsuperscript{1} \\
        Jingxiang Qu\textsuperscript{1} ~~
        Shengfeng He\textsuperscript{2} ~~
        Wenqi Ren\textsuperscript{3} \\
        \textsuperscript{1}Wuhan University of Technology ~~ \textsuperscript{2}Singapore Management University \\
        \textsuperscript{3}Sun Yat-sen University
		% For a paper whose authors are all at the same institution,
		% omit the following lines up until the closing ``}''.
		% Additional authors and addresses can be added with ``\and'',
		% just like the second author.
		% To save space, use either the email address or home page, not both
		% \and
		% Yuan Gao\\
		% Institution2\\
		% First line of institution2 address\\
	}
	\maketitle
	
	%%%%%%%%% ABSTRACT
	\begin{abstract}
		The presence of non-homogeneous haze can cause scene blurring, color distortion, low contrast, and other degradations that obscure texture details. Existing homogeneous dehazing methods struggle to handle the non-uniform distribution of haze in a robust manner. The crucial challenge of non-homogeneous dehazing is to effectively extract the non-uniform distribution features and reconstruct the details of hazy areas with high quality. In this paper, we propose a novel self-paced semi-curricular attention network, called SCANet, for non-homogeneous image dehazing that focuses on enhancing haze-occluded regions. Our approach consists of an attention generator network and a scene reconstruction network. We use the luminance differences of images to restrict the attention map and introduce a self-paced semi-curricular learning strategy to reduce learning ambiguity in the early stages of training. Extensive quantitative and qualitative experiments demonstrate that our SCANet outperforms many state-of-the-art methods. The code is publicly available at \url{https://github.com/gy65896/SCANet}.
	\end{abstract}
	
	%%%%%%%%% BODY TEXT
	\section{Introduction}\label{sec: intro}
	\begin{figure}[t]
		\centering
		\includegraphics[width=1\linewidth]{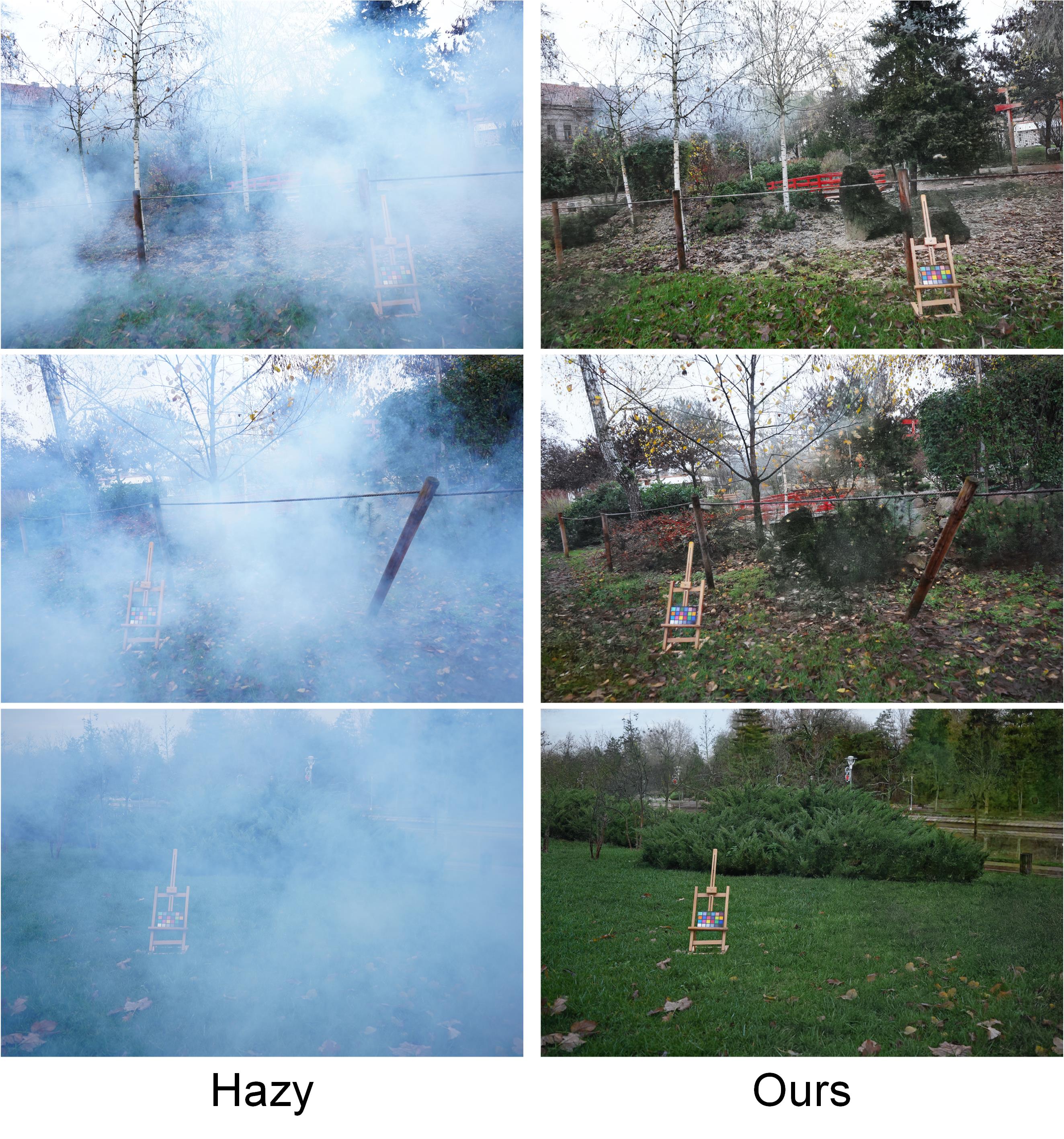}
		\vspace{-8mm}\caption{Dehazing results of the proposed SCANet on the NTIRE2023 test set. Our method can reconstruct high-quality haze-free images.}
		\label{fig:performance}
	\end{figure}
     The existence of turbid media in the atmosphere can lead to the absorption and scattering of light, resulting in degraded hazy scenes that adversely affect the performance of vision-driven scene understanding and object detection methods \cite{zhou2022mtanet, yang2022real}. To tackle this issue, many physical prior-based image dehazing models have been proposed \cite{he2010single, huang2014efficient, fattal2014dehazing, zhu2015fast, berman2016non, zhu2018haze, shu2019variational}. These models typically represent the imaging process using an atmospheric scattering model, which can be expressed as follows
	\begin{equation}
	I(x)=J(x)t(x)+A(x)(1-t(x)),
	\end{equation}
	where $x$ is the pixel index, $I$, $J$, $t$, and $A$ represent the hazy image, clear image, transmission map, and global atmospheric light, respectively. However, it is critical for the success of physical prior-based dehazing methods to estimate $t$ and $A$. When hazy scenes are complex, the estimation of $t$ and $A$ may be inaccurate, leading to unsatisfactory dehazing performance. To achieve superior dehazing performance, numerous learning-based single image dehazing methods \cite{cai2016dehazenet, li2017aod, li2018single, chen2019gated, zhang2019famed, dudhane2019cdnet, qin2020ffa, liu2020trident, mehta2020hidegan, dong2020multi, hong2020distilling, fu2021dw, liu2022deep} have been proposed by leveraging the powerful nonlinear feature representation capacity of deep neural networks. However, haze may be spatially variable and non-uniform in the realistic scenes, making many physical prior- and learning-based methods designed for homogeneous haze inapplicable.

\begin{figure}[t]
		\centering
		\includegraphics[width=1\linewidth]{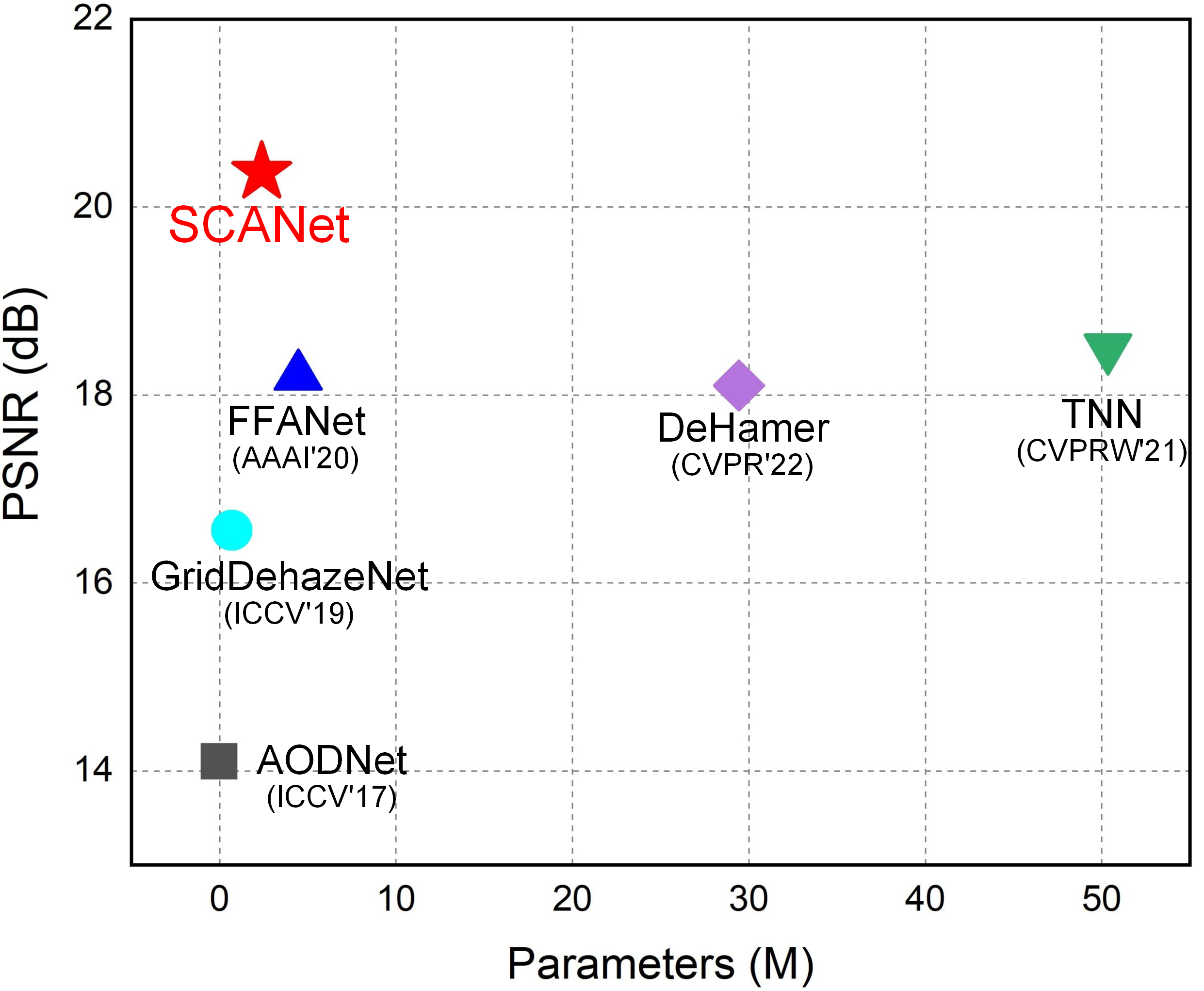}
		\caption{Comparisons of PSNR and parameters of several state-of-the-art dehazing methods on 15 non-homogeneous images from NTIRE2020, NTIRE2021, and NTIRE2023 datasets.}
		\label{fig:metric}
	\end{figure}

 	In recent years, many methods have been proposed to address the challenge of non-homogeneous image dehazing \cite{jo2021multi, yu2021two, shetty2023non,zheng2023curricular}. However, modeling the complex interactions between non-homogeneous haze and the underlying scene remains a challenging task. The key challenge is accurately perceiving the distribution of haze and reconstructing the texture detail of haze-dense areas with high quality. To address this issue, we propose a self-paced semi-curricular attention network (SCANet) for non-homogeneous image dehazing, which consists of an attention generation network and a scene reconstruction network. To better restore areas with significant luminance changes, we design a self-paced semi-curricular learning strategy to control the generation of attention maps. Figure \ref{fig:performance} displays three dehazing cases on the NTIRE2023 test set. The proposed SCANet can adaptively extract non-homogeneous haze features and effectively suppress its interference. Furthermore, Figure \ref{fig:metric} compares the peak signal-to-noise ratio (PSNR) and parameters of our method with state-of-the-art methods, demonstrating the competitive performance of our SCANet.

  	Overall, our main contributions are as follows
	\begin{itemize}
		\item To address the challenging problem of non-homogeneous image dehazing, we propose an attention network that learns complex interaction features between non-homogeneous haze and the underlying scene. The proposed method employs a novel ``attention generation-scene reconstruction'' paradigm specifically designed for non-homogeneous image dehazing.
		\item To enhance the haze removal ability in areas with significant luminance differences, we introduce a self-paced semi-curricular learning-driven attention map generation strategy. This approach improves model convergence and reduces the learning ambiguity caused by multi-objective prediction in the early stages of training.
		\item We extensively evaluate the proposed SCANet through qualitative and quantitative experiments, demonstrating its superior performance compared to state-of-the-art methods. We conduct an ablation analysis to confirm the effectiveness of our method, highlighting the contribution of each component to the overall performance of SCANet.

	\end{itemize}
	\begin{figure*}[t]
		\centering
		\includegraphics[width=1\linewidth]{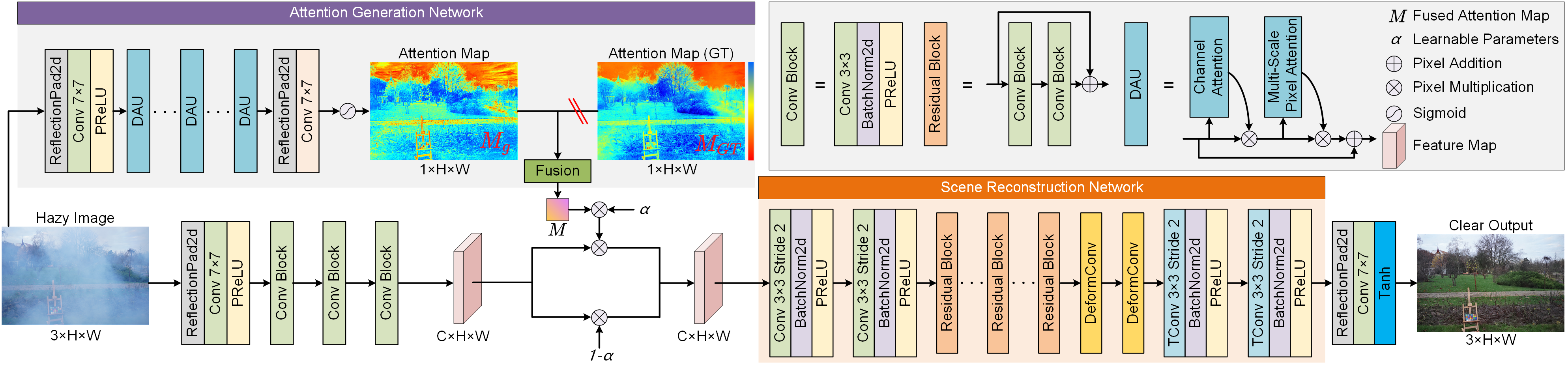}
		\caption{The network structure of our SCANet. The proposed method comprises an attention generation network and a scene reconstruction network. The red slash means we only use $M_{GT}$ during the training phase.}
		\label{fig:scanet}
	\end{figure*}

	\section{Related Works}\label{sec: Related}
	\textbf{Physical Prior-Based Dehazing.} Physical prior-based methods depend on the physical scattering model. Some methods treat empirical observation as the prior knowledge to restore a hazy image, such as dark channel prior (DCP) \cite{he2010single}, color attenuation prior \cite{zhu2014single}, and non-local prior \cite{berman2016non}. He \textit{et al.} \cite{he2010single} proposed a dark channel prior (DCP) of clean outdoor images in terms of pixel intensities and achieved a nice dehazing performance. Zhu \textit{et al.} \cite{zhu2014single} discovered the brightness and saturation of the pixels in hazy images are different and proposed the color attenuation prior. Berman \textit{et al.} \cite{berman2016non} proposed an effective non-local path prior based on the observation that the pixel are usually non-local in a given RGB space. While these priors can yield impressive results in certain scenarios, they may not always be practically applicable. In the real world, haze is often influenced by a variety of complex factors, making these priors unsuitable and resulting in suboptimal dehazing outcomes. For instance, the DCP \cite{he2010single} fails to dehaze the sky regions properly due to the inapplicable prior assumption.
	\textbf{Deep Learning-Based Dehazing.} With the rapid advancement of deep learning, numerous learning-based dehazing methods have been proposed. Cai \textit{et al.} \cite{cai2016dehazenet} introduced an end-to-end network (DehazeNet), which generates the transmission map of the hazy image and recovers a clear image via the atmospheric scattering model. Li \textit{et al.} \cite{li2017aod} proposed  an all-in-one dehazing network (AODNet) that jointly estimates the atmospheric light and transmittance to recover the hazy image. Ren \textit{et al.} \cite{ren2018gated} applied a fusion-based strategy using a multi-scale structure in their haze-free image generation framework. Zhang \textit{et al.} \cite{zhang2018densely} proposed a densely connected pyramid dehazing network (DCPDN), which estimates the transmission map using an edge-preserving densely connected en-decoder structure with a multilevel pyramid pooling module. Qu \textit{et al.} \cite{qu2019enhanced} proposed an enhanced pix2pix dehazing network (EPDN) that uses a generative adversarial network and an enhancer to accomplish the dehazing task. Chen \textit{et al.} \cite{chen2019gated} introduced a gated context aggregation network (GCANet) that uses the smoothed dilation technique to efficiently generate a haze-free image. Liu \textit{et al.} \cite{liu2019griddehazenet} proposed an attention-based multi-scale network (GridDehazeNet), which learns the feature map directly instead of estimating the transmission map. Recently, some studies \cite{chen2019gated, deng2020hardgan, qin2020ffa, wu2021contrastive} tend to estimate the haze-free image or the residual between the hazy image and the corresponding clear image. Hong \textit{et al.} \cite{hong2022uncertainty} proposed an uncertainty-driven dehazing network (UDN) that improves the dehazing results by using the relationship between uncertain and confident representations. Although significant progress has been made by these methods in dehazing tasks, they tend to overlook the issue of non-homogeneous haze suppression. In recent years, several methods \cite{jo2021multi, yu2021two, shetty2023non,zheng2023curricular} have been proposed to address this challenge. However, researchers are still struggling with the difficulty of learning haze distribution features and the poor quality of detail recovery in heavily hazy regions.

	\section{Proposed Method}\label{sec: Proposed Method}
	In this section, we first introduce the network architecture of our SCANet. Then, we describe the proposed self-paced semi-curricular learning-driven attention map generation method. Finally, the loss functions employed in model training are mentioned.
	\subsection{Network Architecture}\label{Network Architecture}
	As illustrated in Figure \ref{fig:scanet}, our method comprises two sub-networks: the attention generation network (AGN) and the scene reconstruction network (SRN). The AGN is composed of multiple dual-attention basic units (DAUs) to generate attention feature maps, while the SRN is an encoder-decoder network to reconstruct haze-free images. %To alleviate the challenge of the task and more focus on the learning of model parameters in the SRN during the early training stage, we adopt a self-paced semi-curriculum learning strategy for generating the attention map, as inspired by \cite{du2023dsdnet}. {\color{red}Our ablation study has demonstrated that the feature map generated by our method can provide better guidance for reconstructing sharp images than unsupervised attention maps.}
	\begin{figure}[t]
		\centering
		\includegraphics[width=1\linewidth]{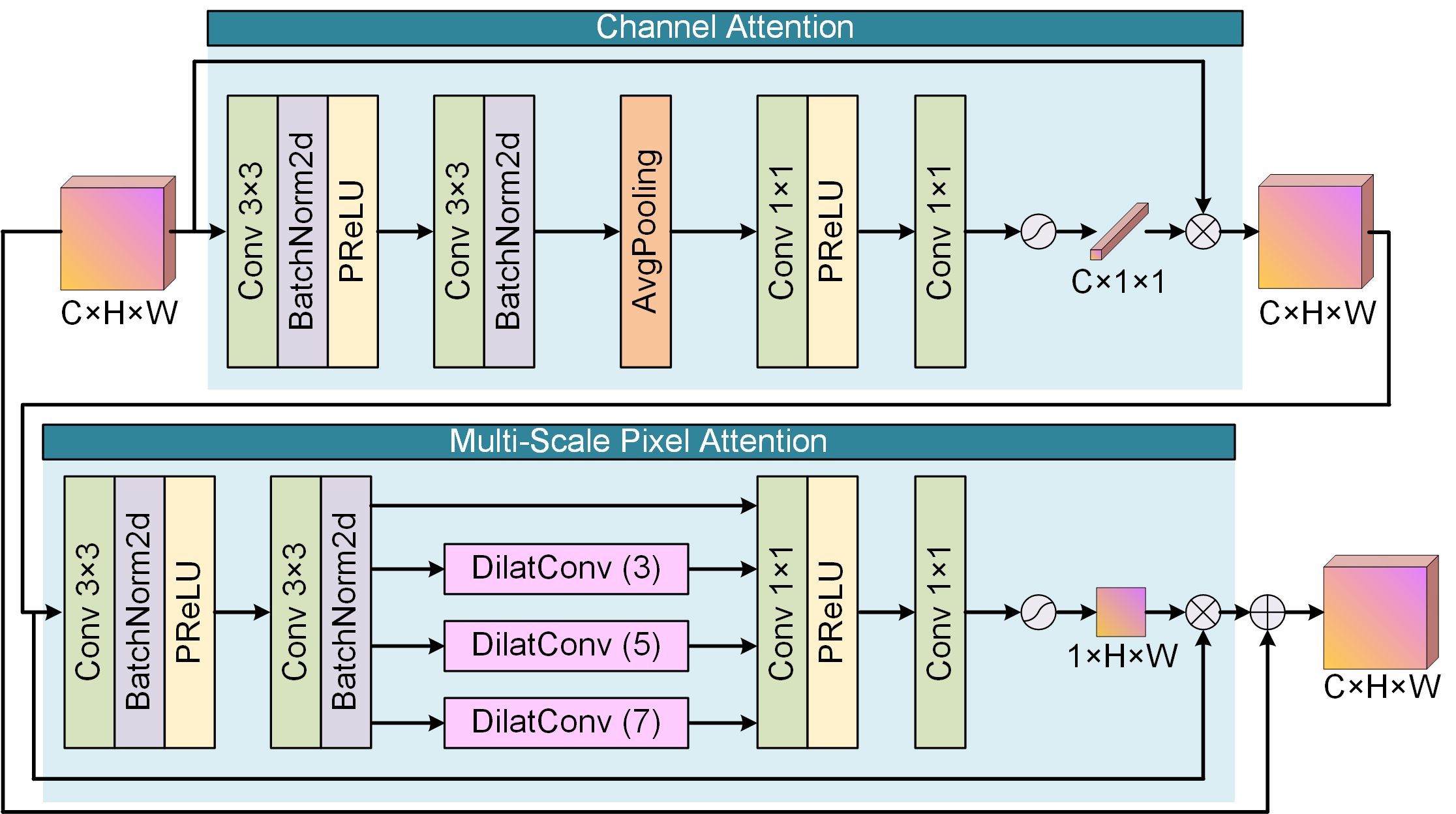}
		\caption{The pipeline of the dual attention unit (DAU). The DAU contains channel attention and multi-scale pixel attention.}
		\label{fig:dau}
	\end{figure}
	\textbf{Attention Generator Network.} Our first sub-network (AGN) is designed to produce the attention feature map. Essentially, the AGN is stacked by multiple dual-attention units (DAUs), as shown in Figure \ref{fig:dau}. The input feature map will be sequentially processed by channel attention (CA) and multi-scale pixel attention (MSPA) to obtain the output feature map. The CA comprises two $3 \times 3$ convolutional layers, a global average pooling layer, two $1 \times 1$ convolutional layers, and a sigmoid function. The obtained weights of each channel by CA will be multiplied by the input feature map. The MSPA includes two $3 \times 3$ convolutional layers, three dilated convolutional layers with different dilated ratios $\in \{3, 5, 7\}$, two $1 \times 1$ convolutional layers, and a sigmoid function. To improve the perception of the spatial distribution of haze, three dilated convolutions are specially used to obtain feature information of multiple receptive fields. Finally, a $7 \times 7$ convolutional layer and a sigmoid function are employed to obtain the attention map $M_g$.

	\textbf{Scene Reconstruction Network.} To improve the haze-free image reconstruction quality, an encoder-decoder network is employed. As illustrated in Figure \ref{fig:scanet}, the SRN first adopts two $3 \times 3$ convolutional layers with the stride of 2 to extract $4\times$ downsampled features. Then, multiple residual blocks and two deformable convolutional layers are used to learn the hazy feature representations in the low-resolution. In particular, the deformable convolution \cite{dai2017deformable} can adjust the kernel shape to focus on the features of interest by using offsets. Subsequently, two transposed convolutional layers with a stride of 2 are used to restore the features to the original resolution. Finally, the haze-free results are produced by a tail block, which contains a reflection padding, a $7 \times 7$ convolutional layer, and a tanh function.
	\subsection{Self-Paced Semi-Curricular Attention}
\begin{figure}[t]
		\centering
		\includegraphics[width=1\linewidth]{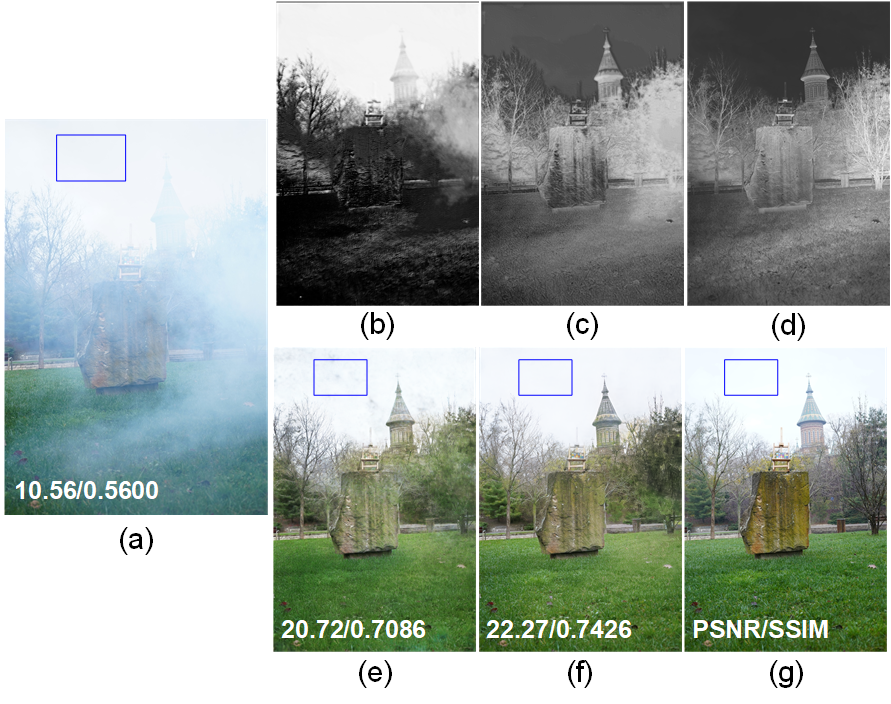}
		\caption{Visual comparisons of images generated by different strategies. From top-left to bottom-right: (a) hazy image, (b) attention map directly generated by AGN, (c) attention map generated by self-paced semi-curricular learning-driven AGN, (d) ground truth of the attention map, (e) dehazing result generated based on (b), (f) dehazing result generated based on (c), and (g) haze-free image. Note that the dehazing result (e) appears over-enhanced and exhibits noticeable artifacts, which can be attributed to the significant weight placed on the sky area by the attention map (b).}
		\label{fig:sca}
	\end{figure}

 	\textbf{Why Supervise Attention Map.} In non-homogeneous image dehazing, attention mechanisms can enable the network to flexibly focus on haze features to reconstruct high-quality haze-free images. However, attention maps are often unsupervised, which can lead to low-importance regions being assigned higher weights and generating low-quality reconstruction results. Figure \ref{fig:sca} (b) and (e) display the attention map directly generated by AGN and the haze-free outputs generated by SRN. Obviously, the attention map has excessively high weights in the sky area, resulting in obvious block artifacts in the reconstruction result. According to our observations, non-homogeneous haze can significantly increase the luminance of occluded areas (except for the sky area). Theoretically, paying more attention to the restoration of areas with significant luminance changes can avoid the over-enhancement issue to improve the overall image reconstruction performance. Therefore, we transform the hazy and clear images into the YCbCr color space and calculate the Y channel-based luminance deviation as the ground truth of the attention map $M_{GT}$.

 	\textbf{Self-Paced Semi-Curricular Learning.} Note that multi-objective prediction tasks (i.e., obtaining both haze-free image and attention map) tend to increase the learning ambiguity. To make the model converge better, inspired by \cite{du2023dsdnet}, we adopt a self-paced semi-curricular learning strategy to train the network from easy to hard. During training, the attention map $M_{g}$ generated by AGN and the ground truth $M_{GT}$ are fused to generate the final attention map $M$. Let $\lambda$ be the trade-off parameter, $M$ can be expressed mathematically as
        \begin{equation}
	       M = \lambda \cdot M_g + (1-\lambda) \cdot M_{GT}.
	\end{equation}

 	In particular, the trade-off parameter can be dynamically adjusted through the smooth L1 loss $\mathcal{L}^{a}_{sl1}$ of the attention map, i.e.,

        \begin{equation}\label{eq:sca}
	       \lambda= \begin{cases}0, & \text { if } \mathcal{L}^{a}_{sl1} > 0.1, \\ \frac{\mathcal{L}^{a}_{sl1}-0.1}{0.1-0.05}, & \text { if } 0.1 \geq \mathcal{L}^{a}_{sl1} > 0.05, \\ 1, & \text { if } \mathcal{L}^{a}_{sl1} \leq 0.05.\end{cases}
	\end{equation}
 
 	Eq. (\ref{eq:sca}) is used to adjust the specific gravity of $M_g$ and $M_{GT}$. In the initial stage, $M$ mainly consists of $M_{GT}$ to alleviate the learning ambiguity due to the large value of $\mathcal{L}^{a}_{sl1}$. As $\mathcal{L}^{a}_{sl1}$ decreases, the proportion of the attention map $M_g$ generated by the network will continue to increase. When $\mathcal{L}^{a}_{sl1}$ is less than 0.05, $M$ will only consist of $M_g$. Meanwhile, we only adopt the semi-curricular learning strategy in the first 25\% epochs to avoid the model’s over-reliance on $M_{GT}$.

 	After obtaining the attention map $M$, we adaptively weight the feature map through a learnable parameter $\alpha$. Let $F_{in}$ be the input feature map, the feature map $F_{out}$ weighted by the attention map can be given by
        \begin{equation}
	       F_{out} = (1 - \alpha) \cdot F_{in} + \alpha \cdot M \otimes F_{in},
	\end{equation}
   	with $\otimes$ being the operator of pixel-wise multiplication.
	\subsection{Loss function}\label{loss}
	In this section, we introduce the joint loss function of the proposed SCANet. Specifically, this joint loss function $\mathcal{L}_{\text {joint}}$ mainly consists of smooth L1 loss  (including $\mathcal{L}_{s l 1}$ and $\mathcal{L}_{s l 1}^a$), multi-scale structural similarity (MS-SSIM) loss $\mathcal{L}_{\text {MS-SSIM}}$, perceptual loss $\mathcal{L}_{\text {p}}$, and adversarial loss $\mathcal{L}_{\text {a}}$, which can be expressed as follows
	\begin{equation}
	       \mathcal{L}_{\text {joint}}=\gamma_1\mathcal{L}_{s l 1}+
	       \gamma_2\mathcal{L}_{s l 1}^a+
	       \gamma_3\mathcal{L}_{p}+
	       \gamma_4\mathcal{L}_{\text {MS-SSIM}}+
	       \gamma_5\mathcal{L}_{a},
	\end{equation}
	where $\gamma_1$, $\gamma_2$, $\gamma_3$, $\gamma_4$, and $\gamma_5$ are the hyper-parameters. The best performance is achieved when we assign them the values of 1, 0.3, 0.01, 0.5, and 0.0005, respectively.
 %In our experiment, giving them the value of 1, 0.3, 0.01, 0.5, and 0.0005 achieves the best performance.
	%
	%

	\textbf{Smooth L1 Loss.} In the image restoration task, Zhao \emph{et al.} \cite{zhao2016loss} have demonstrated that the L1 loss function has better effects compared with  L2 loss. Therefore, we use smooth L1 loss \cite{girshick2015fast} to supervise the final output $\hat{J}$ and the predicted attention map $M_{g}$, which can be expressed as follows
	\begin{equation}
	       \mathcal{L}_{s l 1}=L_1(\hat{J}-J),
	\end{equation}
	\begin{equation}
	       \mathcal{L}_{s l 1}^a=L_1(M_{g}-M_{GT}),
	\end{equation}
	where $L_1\left(\cdot\right)$ represents the smooth L1 loss function, $\mathcal{L}_{s l 1}$ is the loss between the network's output $\hat{J}$ and the ground truth $J$, $\mathcal{L}_{s l 1}^a$ is the loss between the predicted attention map $M_{g}$ and the ground truth of attention map $M_{GT}$.
	Let $Q$ denotes the input, the $L_1$ operation can be expressed as follows
	\begin{equation}
	       L_1(Q)=\frac{1}{N} \sum_{i=1}^N \mathcal{D}_{l 1}(Q(i)),
	\end{equation}
	where $i$ is the index pixel, $N$ denotes the sum of pixels. Finally, the smooth L1 operator $\mathcal{D}_{l 1}$ can be given by
	\begin{equation}
	       \mathcal{D}_{l 1}(Q(i))= \begin{cases}0.5 \cdot Q^2(i), & \text { if }|Q(i)|<1, \\ |Q(i)|-0.5, & \text { otherwise. }\end{cases}
	\end{equation}

	\textbf{Perceptual Loss.} To improve the similarity between the output and ground truth in feature space, we add the perceptual loss $\mathcal{L}_{\text {p}}$, which can be written as follows
        \begin{equation}
            \mathcal{L}_{\text {p}}=\frac{1}{3} \sum_r \frac{\left\|\phi_J^{v}\left(\hat{J}\right)-\phi_k^{v}\left(J\right)\right\|_2^2}{C_k H_k W_k},
        \end{equation}
        where $\phi_k^{v}(\cdot)$ represents the feature map of VGG16 in $k$-layer, and $\left(C_k, H_k, W_k\right)$ denotes the shape of the feature map in the corresponding layer. In this paper, $r\in\{$ relu1\_2, relu2\_2, relu3\_3\}.

	\textbf{MS-SSIM Loss.} To improve the contrast of high-frequency regions in the image, we adopt MS-SSIM loss $\mathcal{L}_\text{MS-SSIM}$, which can be defined as follows
	\begin{equation}
	\begin{aligned}
	       \mathcal{L}_\text{MS-SSIM}= & L_\text{MS-SSIM}(J, \hat{J}),
	\end{aligned}
	\end{equation}
	where $L_\text{MS-SSIM}(\cdot)$ represents the multi-scale structure similarity function. The SSIM value can be written as follows
	\begin{equation}
	\begin{aligned}
	       \operatorname{SSIM}(x) & =\frac{2 \mu_J \mu_{\hat{J}}+c}{\mu_J^2+\mu_{\hat{J}}^2+c} \cdot \frac{2 \sigma_{J \hat{J}}+c_*}{\sigma_J^2+\sigma_{\hat{J}}^2+c_*} \\
	& =l(x) \cdot {cs}(x),
	\end{aligned}
	\end{equation}
	where $x$ demotes the pixel index, $c$ and $c_*$ are two constants to avoid the denominator becoming zero. The means $\mu_J$ , $\mu_{\hat{J}}$, standard deviations $\sigma_J$, $\sigma_{\hat{J}}$, and covariance $\sigma_{J \hat{J}}$ are computed by a Gaussian filter. Finally, the operation of MS-SSIM can be defined as follows
	\begin{equation}
	       L_\text{MS-SSIM}=1-l_\mathcal{P}^\alpha \cdot \prod_{j=1}^\mathcal{P}\left[c s_j\right]^{\beta_j},
	\end{equation}
	where $\mathcal{P}$ denotes the default parameter of scales.

	\textbf{Adversarial Loss.} To improve the generalization ability of the proposed network, we add the additional adversarial loss, i.e.,
        \begin{equation}
            \mathcal{L}_{a}=-\frac{1}{S} \sum_{n=1}^S \log \left(D\left(J-\hat{J}\right)\right),
        \end{equation}
	where $D(\cdot)$ represents the discriminator, $S$ represents the number of training data.
	\setlength{\tabcolsep}{2.5pt}
	\begin{table}[t]
		\centering
		\caption{The details of the datasets used in our experiments. (w/o GT) represents the lack of public ground truth for this set.}
            \begin{tabular}{ccccc}
                \hline
                Datasets  & Train & Validation & Test & Image Size         \\\hline\hline
                NTIRE2020 & 45    & 5          & 5    & 1200 $\times$ 1600 \\
                NTIRE2021 & 25    & 5 (w/o GT)          & 5 (w/o GT)    & 1200 $\times$ 1600 \\
                NTIRE2023 & 40    & 5 (w/o GT)          & 5 (w/o GT)    & 4000 $\times$ 6000          \\ \hline
		\end{tabular}\label{table:dataset}
	\end{table}

	\setlength{\tabcolsep}{6pt}
	\begin{table*}[t]
		\centering
		\caption{Quantitative comparisons for non-homogeneous dehazing on NHIRE2020, NHIRE2021, and NHIRE2023 datasets. The best results are in \textbf{bold}, and the second best are with \underline{underline}.}
            \begin{tabular}{lcccccccc}
                \hline
                \multirow{2}{*}{Methods} & \multicolumn{2}{c}{NTIRE2020} & \multicolumn{2}{c}{NTIRE2021} & \multicolumn{2}{c}{NTIRE2023} & \multicolumn{2}{c}{Average} \\
                & PSNR $\uparrow$ & SSIM $\uparrow$ & PSNR $\uparrow$ & SSIM $\uparrow$ & PSNR $\uparrow$ & SSIM $\uparrow$ & PSNR $\uparrow$ & SSIM $\uparrow$ \\ \hline\hline
                Hazy & 11.31 & 0.4160 & 11.24 & 0.5787 & 8.86 & 0.4702 & 10.47 & 0.4883 \\
                (TPAMI'10) DCP   \cite{he2010single} & 12.35 & 0.4480 & 10.57 & 0.6030 & 10.98 & 0.4777 & 11.30 & 0.5096 \\
                (ICCV'17) AODNet   \cite{li2017aod} & 14.04 & 0.4450 & 14.52 & 0.6740 & 13.75 & 0.5619 & 14.10 & 0.5603 \\
                (ICCV'19) GridDehazeNet   \cite{liu2019griddehazenet} & 14.78 & 0.5074 & 18.05 & 0.7433 & 16.85 & 0.6075 & 16.56 & 0.6194 \\
                (AAAI'20) FFANet   \cite{qin2020ffa} & 16.98 & 0.6105 & 19.75 & \underline{0.7925} & 17.85 & \underline{0.6485} & 18.20 & 0.6838 \\
                (CVPRW'21) TNN   \cite{yu2021two} & 17.18 & 0.6114 & \underline{ 20.13} & \textbf{0.8019} & \underline{18.19} & 0.6426 & \underline{18.50} & \underline{0.6853} \\
                (CVPR'22) DeHamer   \cite{guo2022image} & \underline{18.53} & \underline{0.6201} & 18.17 & 0.7677 & 17.61 & 0.6051 & 18.10 & 0.6693 \\\hline
                SCANet & \textbf{19.52} & \textbf{0.6488} & \textbf{21.14} & 0.7694 & \textbf{20.44} & \textbf{0.6616} & \textbf{20.37} & \textbf{0.6933}         \\ \hline
		\end{tabular}\label{table:ntire20-23}
	\end{table*}
        \section{Experiments}\label{sec: Experiments}
        \setlength{\tabcolsep}{4.5pt}
        \begin{table}[t]
            \caption{FLOPs and Parameters comparisons of all methods.}
            \label{tab:my-table}
            \begin{tabular}{lcc}
            \hline
            Methods                & FLOPs  & Parameters \\ \hline \hline 
            (ICCV'17) AODNet \cite{li2017aod}        & 1.68G    & 1.76K       \\
            (ICCV'19) GridDehazeNet \cite{liu2019griddehazenet} & 271.95G  & 702.47K       \\
            (AAAI'20) FFANet \cite{qin2020ffa}        & 4211.91G  & 4.46M       \\
            (CVPRW'21) TNN \cite{yu2021two}          & 1235.84G & 50.35M      \\
            (CVPR'22) DeHamer \cite{guo2022image}       & 866.96G  & 29.44M      \\ \hline
            SCANet                 & 258.63G  & 2.39M       \\ \hline
            \end{tabular}
            \label{table:complex}
        \end{table}
	In this section, we first describe the datasets, implementation details, evaluation metrics, and competitors. Then, we compare the proposed SCANet with other state-of-the-art dehazing methods. Finally, we conduct the ablation study to demonstrate the rationality of each module in the proposed SCANet. 
	\subsection{Experiment Settings}
	\textbf{Datasets.} We choose NTIRE2020 \cite{ancuti2020nh, ancuti2020ntire}, NTIRE2021 \cite{ancuti2021ntire}, and NTIRE2023 \cite{ancuti2023ntire} datasets to train and evaluate the proposed SCANet. The haze patterns in all three datasets are non-uniformly distributed. Specifically, NTIRE2020 dataset (termed NH-Haze) contains 45 training, 5 validation, and 5 test image pairs. NTIRE2021 dataset (termed NH-Haze2) contains 25 training image pairs, 5 validation hazy images, and 5 test hazy images. NTIRE2023 dataset contains 40 training image pairs, 5 validation hazy images, and 5 test hazy images. Note that only the validation and test sets of NTIRE2020 dataset contain the corresponding ground truth. More details about these datasets can be found in Table \ref{table:dataset}.
	\textbf{Implementation Details.} The proposed SCANet is implemented by PyTorch 1.9.1 and trained on a PC with an Intel(R) Core(TM) i9-13900K CPU @5.80GHz and Nvidia GeForce RTX 3080 GPU. We use the Adam with exponential decay rates being $\beta_1=0.9$ and $\beta_2=0.999$ for optimization. The initial learning rate and batchsize are set to 0.0001 and 2, respectively. During the training stage, we resize the images into 0.5, 0.7, and 1 scales and randomly crop them to several image patches of size 512 $\times$ 512 with a stride of 400. Meanwhile, these image patches are randomly flipped 0, 90, 180, and 270 degrees. In addition, we train two models for NTIRE2023 validation and test sets and NTIRE2020/2021/2023 datasets, respectively. For NTIRE2023 validation and test sets, we only use 35 training pairs in NTIRE2023 for training. The epoch is set to 85, and the learning rate decays by 0.5 every 20 epochs. Due to the large size of the test images, we adopt the Nvidia A100 GPU for testing. For NTIRE2020, NTIRE2021, and NTIRE2023 datasets, we select 45 training pairs and 5 validation pairs in NTIRE2020, the first 20 training pairs in NTIRE2021, and the first 35 training pairs in NTIRE2023 as the train set. The test set is composed of 5 test pairs in NTIRE2020, the last 5 training pairs in NTIRE2021, and the last 5 training pairs in NTIRE2023. In this experiment, the images of NTIRE2023 are compressed to 1/4 (i.e., 1000 $\times$ 1500) to ensure a similar size with other datasets. In addition, the epoch is set to 500, and the learning rate decays by 0.5 every 150 epochs.

 	\textbf{Evaluation Metrics and Competitors.} To conduct an exhaustive analysis of the dehazing performance, we employ the peak signal-to-noise ratio (PSNR) \cite{wang2009mean} and structural similarity index (SSIM) \cite{wang2004image} to quantitatively evaluate the restored images. Meanwhile, we compare the proposed SCANet with the state-of-the-art methods, including a prior-based method (i.e., DCP \cite{he2010single}), a physical model-based CNN method (i.e., AODNet \cite{li2017aod}), three hazy-to-clear CNN methods (i.e., GridDehazeNet \cite{liu2019griddehazenet}, FFANet \cite{qin2020ffa}, and TNN \cite{yu2021two}), and a CNN-Transoformer combined method (i.e., DeHamer \cite{guo2022image}).

	\subsection{Comparisons with the State-of-the-Arts}
	\begin{figure*}[t]
		\centering
		\includegraphics[width=1\linewidth]{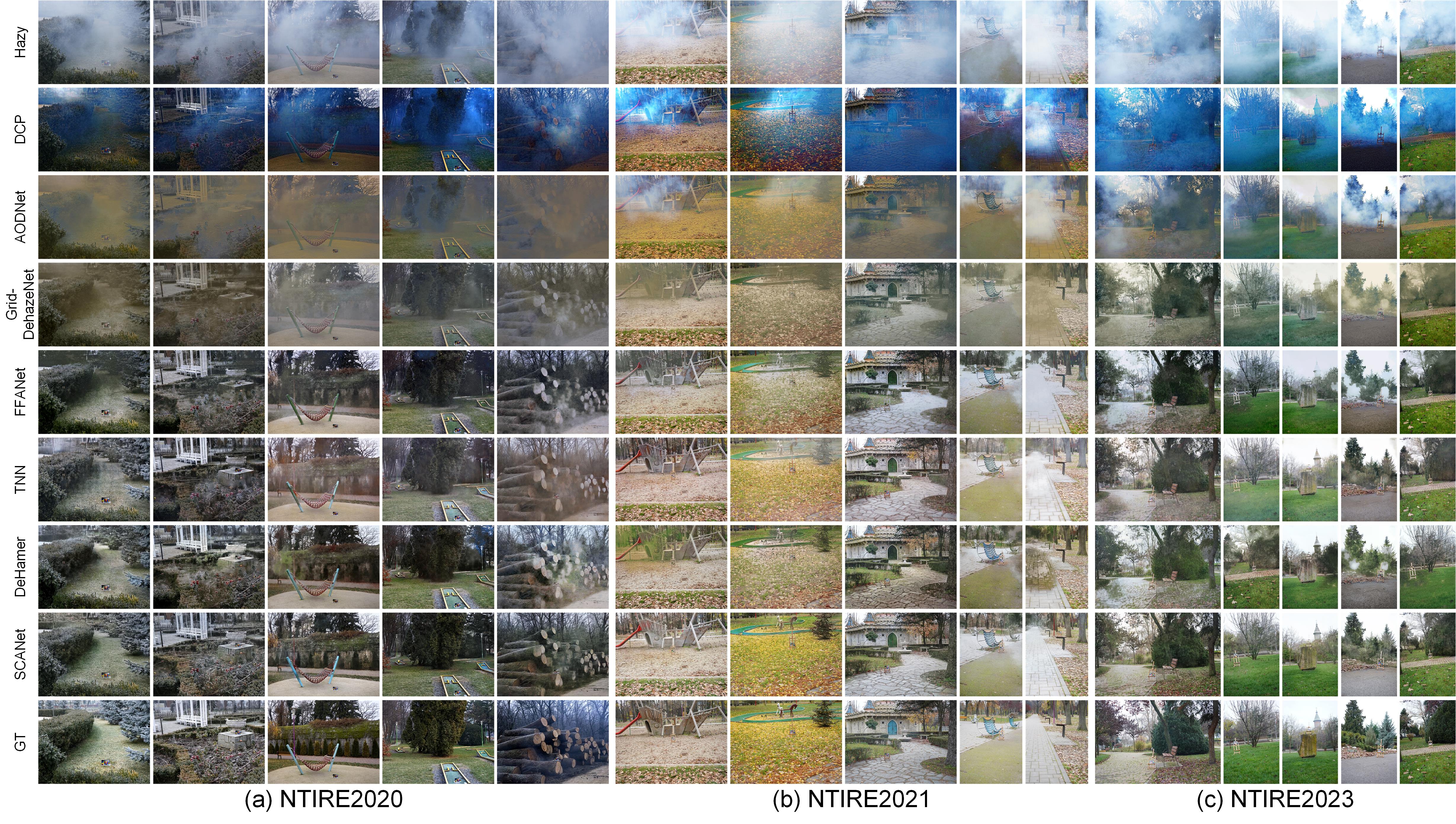}
		\caption{Visual comparisons of various methods on NTIRE2020, NTIRE2021, and NTIRE2023 datasets.}
		\label{fig:ntire20-23}
	\end{figure*}
	\begin{figure*}[t]
		\centering
		\includegraphics[width=1\linewidth]{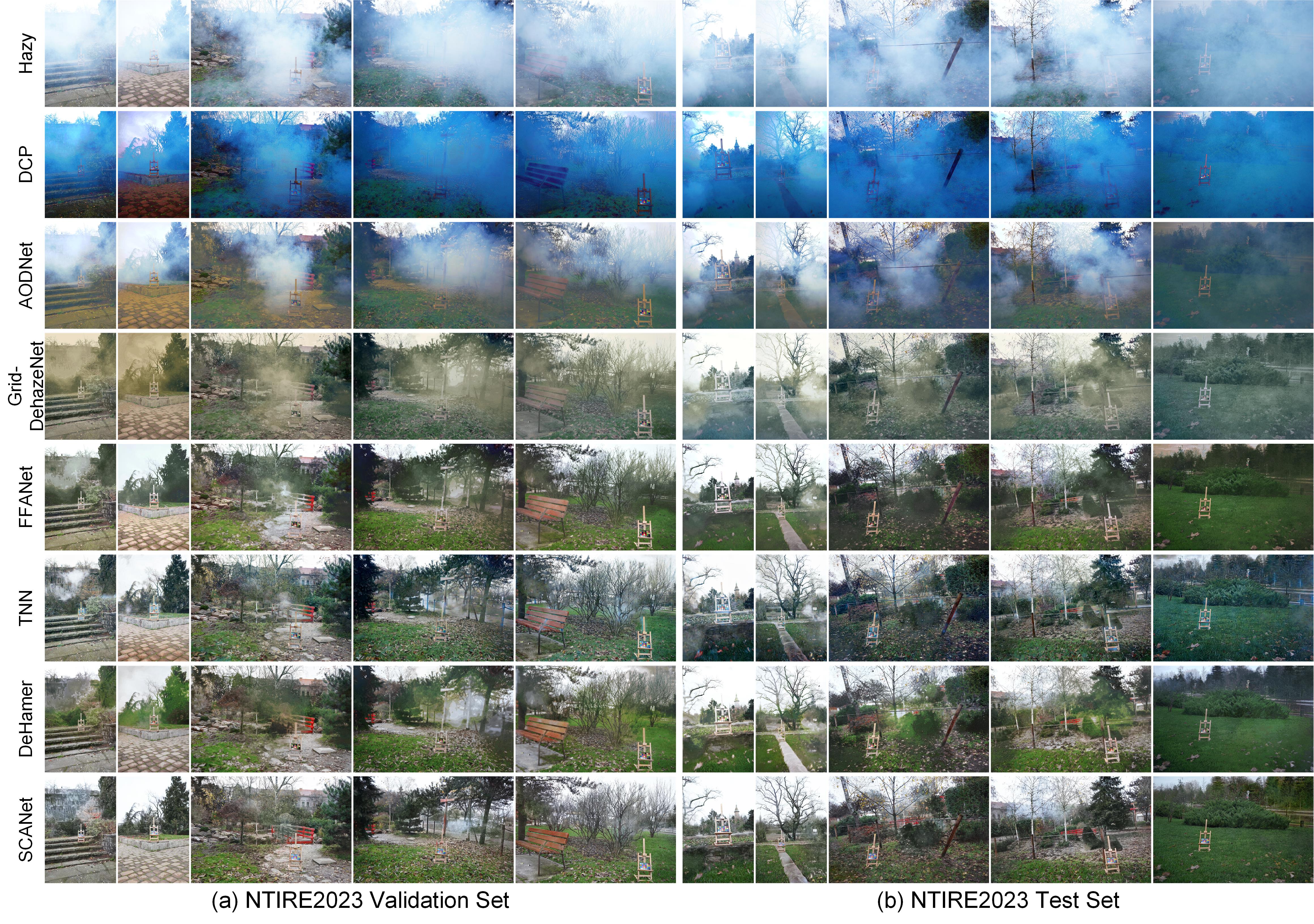}
		\vspace{-8mm}\caption{Visual comparisons of various methods on NTIRE2023 validation set ($\#41\sim45$) and test set ($\#46\sim50$).}
		\label{fig:ntire23}
	\end{figure*}

	\textbf{Results on NTIRE2020/2021/2023.} Table \ref{table:ntire20-23} presents the PSNR and SSIM results of various dehazing methods on NTIRE2020, NTIRE2021, and NTIRE2023 datasets. Prior knowledge of DCP fails in the non-homogeneous dehazing task, resulting in relatively low values of PSNR and SSIM. Learning-based methods show better adaptability in generating haze-free images, with a significant boost in metrics. Among these methods, the proposed SCANet achieves satisfactory performance, ranking first in most cases. We also show the visual comparisons in Figure \ref{fig:ntire20-23}. The results generated by DCP have the issue of serious color distortion. AODNet, GridDehazeNet, and FFANet fail to remove haze completely. The performance of TNN on the NTIRE2023 benchmark falls short of expectations. Specifically, five images lack sufficient color saturation, while the first image exhibits overly darkened ground. DeHamer is effective in haze suppression. However, color restoration and detail preservation abilities still need improvement. Compared with other methods, the proposed SCANet exhibits superior visual performance.

	\textbf{Results on NTIRE2023 Validation and Test Sets.} According to our submission on the NTIRE2023 website, our SCANet can achieve PSNR 21.13dB and SSIM 0.6907 on the validation set and PSNR 21.75dB and SSIM 0.6955 on the test set. Meanwhile, the visual comparison of our method and the state-of-the-art on 5 validation and 5 test images are shown in Figure \ref{fig:ntire23}. It can be observed that DCP, AODNet, and GridDehazeNet perform poorly in non-homogeneous image dehazing. Although FFANet, TNN, and DeHamer can partially remove the haze, residues still exist in dense haze regions. Compared to existing methods, the proposed SCANet have a more natural performance. However, our method still cannot fully restore the color and details of high haze concentration areas.

	\textbf{Complexity Analysis.} Table \ref{table:complex} shows the number of network parameters and floating point operations (FLOPs) on 1200 $\times$ 1600 images of the proposed method and other comparable methods. By comparison, our SCANet has lower FLOPs and fewer network parameters. To visually demonstrate the superiority of our method, we compare the PSNR and parameter amount of different methods in Figure \ref{fig:metric}. It is worth mentioning that our time complexity is also relatively modest. It takes an average of 0.1962 seconds to process a 1200 $\times$ 1600 image on the NVIDIA GeForce RTX 3080 GPU.

	\setlength{\tabcolsep}{11pt}
	\begin{table*}[t]
		\centering
		\caption{The ablation study of different configurations. The best results are in \textbf{bold}, and the second best are with \underline{underline}.}
            \begin{tabular}{ccccccccc}
            \hline
                Number & Methods & $\mathcal{L}_{sl1}^{f}$ & $\mathcal{L}_{sl1}^{a}$ & $\mathcal{L}_{p}$ & $ \mathcal{L}_\text{MS-SSIM}$ & $\mathcal{L}_{a}$ & PSNR $\uparrow$ & SSIM $\uparrow$ \\\hline\hline
                (1) & SRN & \ding{52} &  &  &  &  & 18.84 & 0.6634 \\
                (2) & AGN + SRN & \ding{52} &  &  &  &  & 19.29 & 0.6714 \\
                (3) & SRN + AGN & \ding{52} & \ding{52} &  &  &  & 19.71 & 0.6787 \\
                (4) & SRN + AGN + SCL & \ding{52} & \ding{52} &  &  &  & 19.92 & 0.6881 \\ \hline
                (5) & SRN + AGN + SCL & \ding{52} & \ding{52} & \ding{52} &  &  & 19.85 & 0.6890 \\
                (6) & SRN + AGN + SCL & \ding{52} & \ding{52} & \ding{52} & \ding{52} &  & \underline{20.02} & \textbf{0.6957} \\
                (7) & SRN + AGN + SCL & \ding{52} & \ding{52} & \ding{52} & \ding{52} & \ding{52} & \textbf{20.37} & \underline{0.6933} \\ \hline
		\end{tabular}\label{table:ab}
	\end{table*}

	\subsection{Ablation Analysis}
	We conduct a series of experiments as an ablation study to demonstrate the effectiveness of different components, including attention generation network (AGN), scene reconstruction network (SRN), self-paced semi-curricular learning strategy (SCL), and each loss function. As shown in Table \ref{table:ab}, we design seven models with different configurations and employ NTIRE2020, NTIRE2021, and NTIRE2023 datasets as both training and test sets. 

	The quantitative results are presented in Table \ref{table:ab}. By comparing Model (1) and (2), our method achieves performance improvement after adding the attention generator network (AGN) before the scene reconstruction network (SRN). This result demonstrates that unlike homogeneous image dehazing, restoring non-homogeneous images requires the network to be more sensitive to the haze regions. Moreover, we use $\mathcal{L}_{sl1}^{a}$ to supervise the attention feature map, resulting in satisfactory improvement in both PSNR and SSIM by observing Model (2) and (3). The supervision of the attention map avoids assigning higher weights to low-importance regions, which can provide better reconstruction results. Additionally, applying self-paced semi-curricular learning (SCL) during training leads to the further improvement of metrics, which indicates that SCL can reduce the network's convergence difficulty and improve its performance. By comparing the examples shown in Figure \ref{fig:sca}, we can find the change from Model (2) to Model (4) more intuitively. Obviously, our SCL strategy for attention map constraint can make the SRN more fully focus on the regions with significant luminance changes and avoid the distortion issue in the sky region. Furthermore, the usage of MS-SSIM loss, perceptual loss, and generative adversarial loss can further enhance the dehazing performance of our SCANet by comparing Model (5), (6), and (7) in Table \ref{table:ab}.
 
	\section{Conclusion}
	In this paper, we provided a robust solution (termed SCANet) for non-homogeneous image dehazing by effectively extracting non-uniform haze distribution features and reconstructing the details with high quality. Our attention generator network and scene reconstruction network work together in a novel ``attention generation-scene reconstruction'' paradigm. Moreover, we proposed a self-paced semi-curricular learning-driven attention map generation strategy to improve the model convergence and reduce learning ambiguity during the early stage of training. Our proposed method outperforms many state-of-the-art methods in both quantitative and qualitative experiments, demonstrating the effectiveness of our approach. Additionally, ablation analysis confirms the contribution of each component in the overall performance of our SCANet. We believe that the proposed method can provide a promising solution for the applications of real-world non-homogeneous image dehazing. Future work can extend our method to handle more complex scenarios. Examples include handling multiple types of haze and integrating our method with other computer vision tasks.
        
        \section{Acknowledgements}
        This work is supported by the National Natural Science Foundation of China (No.: 52271365). The authors would like to thank the three anonymous reviewers for their professional comments and constructive suggestions.
	
	%

	%%%%%%%%% REFERENCES
	{\small
		\bibliographystyle{ieee_fullname}
		\bibliography{egbib}
	}
	
\end{document}